\documentclass{article}
\usepackage{spconf,amsmath,graphicx}

\usepackage{enumitem}
\usepackage{amssymb}
\setlist{nosep, leftmargin=14pt}
\usepackage[table,xcdraw]{xcolor}
\usepackage{mwe} 
\usepackage{multirow}  
\usepackage{url} 
\usepackage{hyperref}

\title{Leveraging Persistence Image to Enhance Robustness and Performance in Curvilinear Structure Segmentation\\}
\name{
    \begin{tabular}{@{}c@{}}
    Zhuangzhi Gao\textsuperscript{1,3} \quad
    Feixiang Zhou\textsuperscript{2,3} \quad
    He Zhao\textsuperscript{2} \quad
    Xiuju Chen\textsuperscript{5} \quad
    Xiaoxin Li\textsuperscript{5} \\
    Qinkai Yu\textsuperscript{4,3} \quad
    Yitian Zhao\textsuperscript{6} \quad
    Alena Shantsila\textsuperscript{3} \quad
    Gregory Y.~H.~Lip\textsuperscript{3} \quad
    Eduard Shantsila\textsuperscript{1} \quad
    \textbf{Yalin Zheng}\textsuperscript{2,*}
    \end{tabular}
}

\address{
    $^{1}$Department of Primary Care and Mental Health, University of Liverpool, UK \\
    $^{2}$Department of Eye and Vision Sciences, University of Liverpool, UK \\
    $^{3}$Liverpool Centre for Cardiovascular Science, University of Liverpool, UK \\
    $^{4}$Department of Computer Science, University of Exeter, UK \\
    $^{5}$Xiamen Eye Center, Xiamen University, China \\
    $^{6}$Ningbo Institute of Materials Technology and Engineering, Chinese Academy of Sciences, China \\
    {\small *Corresponding author: \texttt{yalin.zheng@liverpool.ac.uk}}
    \vspace{-0.5cm} 
}

\begin{document}

\maketitle

\begin{abstract}
Segmenting curvilinear structures in medical images is essential for analyzing morphological patterns in clinical applications. Integrating topological properties, such as connectivity, improves segmentation accuracy and consistency. However, extracting and embedding such properties—especially from Persistence Diagrams (PD)—is challenging due to their non-differentiability and computational cost. Existing approaches mostly encode topology through handcrafted loss functions, which generalize poorly across tasks. In this paper, we propose PIs-Regressor, a simple yet effective module that learns persistence image (PI)—finite, differentiable representations of topological features—directly from data. Together with Topology SegNet, which fuses these features in both downsampling and upsampling stages, our framework integrates topology into the network architecture itself rather than auxiliary losses. Unlike existing methods that depend heavily on handcrafted loss functions, our approach directly incorporates topological information into the network structure, leading to more robust segmentation. Our design is flexible and can be seamlessly combined with other topology-based methods to further enhance segmentation performance. Experimental results show that integrating topological features enhances model robustness, effectively handling challenges like overexposure and blurring in medical imaging. Our approach on three curvilinear benchmarks demonstrate state-of-the-art performance in both pixel-level accuracy and topological fidelity. Our code is released at \href{https://github.com/NatsuGao7/TopoUnet.git}{\texttt{https://github.com/NatsuGao7/TopoUnet.git}}
.
\end{abstract}

\section{Introduction}
\label{sec:intro}

Segmentation of curvilinear structures—such as retinal vessels \cite{Zhao2020JointLoss,zhou2025glcp}—is essential for morphological analysis, early diagnosis, and treatment planning in clinical practice, particularly for retinal and cardiovascular diseases (CVD) \cite{ 10.1007/978-3-031-73119-8_8}. Although deep learning has improved segmentation accuracy, pixel-level metrics (e.g., Dice Similarity Coefficient) often overlook structural errors such as broken vessels or missing small branches, which may cause clinical misjudgment. Therefore, a robust model is needed to ensure both pixel-level accuracy and topological consistency across diverse medical scenarios. Topological features capture shape and connectivity and remain invariant under smooth deformations, offering improved stability, generalization, and interpretability for curvilinear segmentation \cite{ferri2019topology,gabella2019topology}.

Topological data analysis (TDA) provides a framework for extracting and understanding the topological features of data~\cite{edelsbrunner2022computational}. Persistent homology (PH), a core technique in TDA, identifies and tracks key topological features such as connected components, loops, and voids on different scales \cite{edelsbrunner2008persistent}. This methodology provides the theoretical foundation for many models that ensure topological consistency, either directly or through variations.
\textbf{(i) PH-Driven Loss Design} Persistence barcodes (PB) and persistence diagrams (PD), as representations of persistent homology, provide a reasonable strategy for constructing loss functions \cite{hu2019topology}. However, due to their discrete and non-differentiable nature, integrating PD and PB into end-to-end networks often requires intricate designs and is computationally intensive. 
\textbf{(ii) Indirect PH Annotations} Another line of work uses indirect proxies of PH, such as fractal feature maps~\cite{huang2025representing} and adjacent anatomy labels~\cite{wang2024avdnet,zhang2023anatomy}, to capture topological cues.
These methods can improve performance but require heavy manual annotation and provide only surrogate representations rather than direct topology encoding.
\textbf{(iii) Indirect PH-Driven Loss Design}
Due to the non-differentiability of PH, several variations of loss functions have been proposed to enforce topological consistency, such as clDice \cite{paetzold2019cldice}, cbDice \cite{shi2024centerline}. However, these methods based on simplicial complex theory do not directly incorporate topological features. They primarily capture connectivity, neglecting many other essential topological characteristics. Moreover, certain processing steps may alter specific details, affecting the precision of the segmentation. For example, during the conversion of the segmentation mask to its skeletonized form, the skeletonization process in clDice \cite{paetzold2019cldice} can introduce errors, potentially compromising the quality of the segmentation.

In this paper, we tackle the challenge of enhancing topological consistency and robustness in curvilinear structure segmentation. Inspired by PI-Net \cite{som2020pi}, which extracts topological features for image classification, we propose PIs-Regressor—a compact network that directly learns an approximation of the PIs—a finite, differentiable representation of topological features—from raw images. Additionally, we further introduce Topology SegNet, which fuses image and topological features to exploit topological cues and improve segmentation. Unlike conventional approaches, we do not rely on complex loss functions, nor do we introduce variants of topological representations. Instead, we directly integrate topological information into the segmentation network, fully leveraging the multifaceted properties of topological features—which extend beyond merely capturing connectivity—to enhance both topological consistency and robustness.

\noindent\textbf{Contribution.} Our main contributions are as follows: \textbf{1:} We propose PIs-Regressor, which efficiently approximates PI from image data. \textbf{2:} We propose a new segmentation network, Topology SegNet, which effectively integrates topological features with image features and applies them to segmentation networks. To the best of our knowledge, this is the first work to incorporate topological features into classical segmentation networks to enhance performance. \textbf{3:} We conduct extensive experiments on curvilinear structure segmentation datasets and show that our method achieves state-of-the-art performance and robustness compared to existing methods that emphasize topology.

\section{Method}
\label{sec:format}
\noindent\textbf{Preliminaries.}
Homology is an algebraic topological property that captures the ``holes'' in a space \cite{hatcher2005algebraic}. These holes, such as connected components for \( k = 0 \) and loops for \( k = 1 \), are represented in the homology group \( k \) -th, denoted \( H_k(X) \), where \( X \) represents a topological space.

\begin{figure}[t]
    \centering
    \includegraphics[width=0.8\columnwidth]{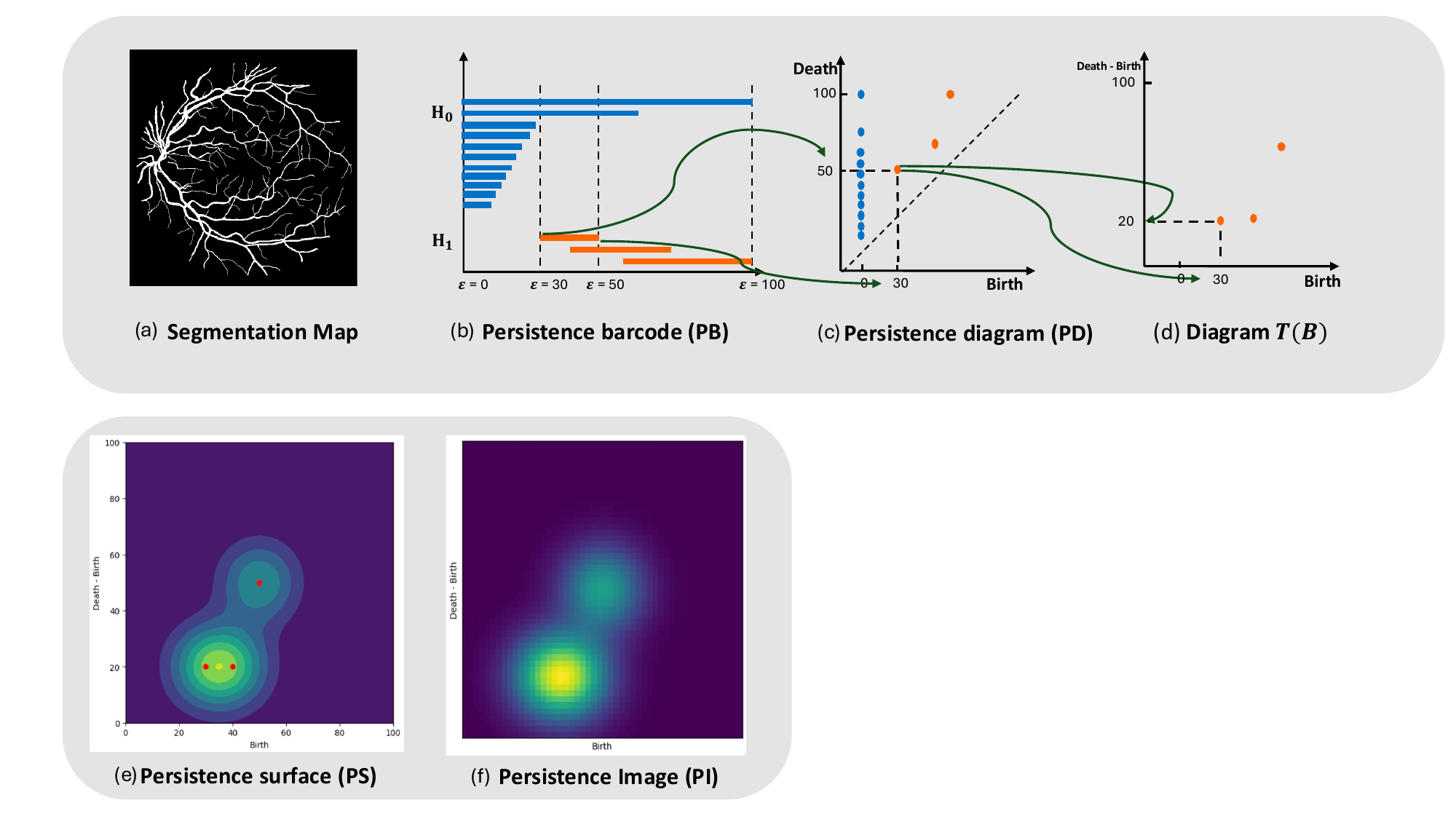}
    \caption{Illustration of the process for obtaining stable topological feature representations—specifically, the persistence image (PI) from segmentation results. The traditional pipeline, as depicted in (a) to (f), often suffers from issues such as non-differentiability and high computational cost.}
    \label{fig:fig1}
\end{figure}

For the curvilinear structure segmentation map of an image, as shown in Fig.\ref{fig:fig1} (a), the foreground can be viewed as a sparse point cloud, denoted as \( P = \{(x, y) \in \mathbb{Z}^2 \mid g(x, y) = 255\} \), representing the vascular structures in the image. As the scale parameter \( \epsilon \) increases, previously disconnected points gradually become connected based on the Euclidean distance \( d(p_1, p_2) \), forming larger connected regions and potential loops. This process results in a nested sequence of topological spaces \( X_1 \subset X_2 \subset \cdots \subset X_n \), with increasing topological complexity. The inclusion \( X_i \subset X_{i'} \) induces a linear map \( H_k(X_i) \to H_k(X_{i'}) \) in homology, allowing persistent homology (PH) to track topological features across scales. In 2D images, 0-dimensional homology \( H_0(X) \) and 1-dimensional homology \( H_1(X) \) are typically used to describe the topology. \( H_1 \) is particularly valuable for identifying vascular structures that persist across scales while filtering out local noise, enabling the retention of biologically meaningful curvilinear structures. Consequently, our analysis focuses on \( H_1 \). A standard way to represent PH is through persistence barcode (PB) and persistence diagram (PD), as shown in Fig. \ref{fig:fig1} (b) and (c). PH effectively distinguishes between persistent structures (e.g., vessel-like shapes) and transient features (e.g., noise or minor artifacts), providing a scale-aware representation of vascular topology.

PD are discrete multisets and cannot be directly used in deep neural networks. To address this, persistence surface in Fig. \ref{fig:fig1} (e) is created by transforming PD points from \( (\text{birth}, \text{death}) \) to \( (\text{birth}, \text{death} - \text{birth}) \) coordinates via \( T(B) \), as shown in Fig. \ref{fig:fig1} (d) for \( H_1 \). The transformed points are weighted by a function \( f(u) \) and a Gaussian distribution \( \phi_u(z) \), yielding a continuous distribution: \( \rho_B(z) = \sum_{u \in T(B)} f(u) \phi_u(z) \), which we refer to as 
persistence surface (PS), as shown in Fig. \ref{fig:fig1} (e). However, PS is still a continuous function, making it unsuitable for direct input into deep neural networks, which typically require fixed-size feature representations. To bridge this gap, we discretize \( \rho_B(z) \) into a structured representation by dividing the domain into \( n \) grid cells (pixels). The PI is then computed by integrating \( \rho_B(z) \) over each cell as \( I(\rho_B)_p = \iint_{p} \rho_B \, dx \, dy \).

This transformation encodes topological features from PD into a fixed-dimensional representation for deep learning while preserving structural and persistence information.

\begin{figure*}[t]
    \centering
    \includegraphics[width=0.9\textwidth]{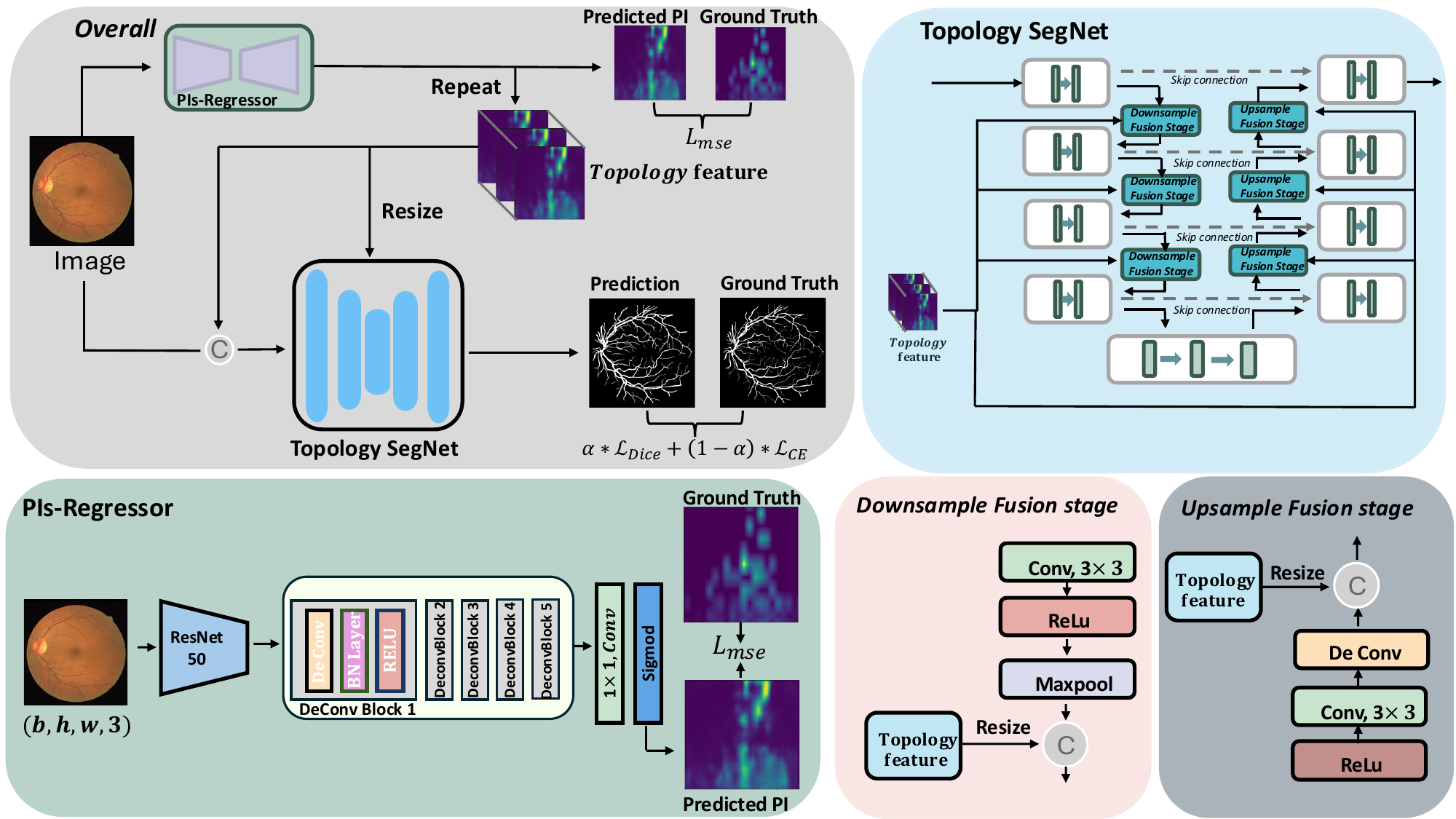}
    \caption{
        Overview of the proposed framework. 
        From top to bottom and left to right: 
        (1) overall model diagram, 
        (2) Topology SegNet architecture, 
        (3) PIs-Regressor network for extracting persistence images, and 
        (4) details of Topology SegNet during downsampling and upsampling
    }
    \label{fig_2}
\end{figure*}

\noindent\textbf{Overall.} As illustrated in Fig.~\ref{fig_2}, the input is an image \( x \in \mathbb{R}^{b \times h \times w \times c} \), where \( b \), \( h \), \( w \), and \( c \) represent the batch size, height, width, and number of channels, respectively. The input image \( x \) is processed by the PIs-Regressor model to obtain an approximation of the PI, denoted as \( p' \in \mathbb{R}^{h \times w \times 1} \), and trained using the Mean Squared Error (MSE) loss function. During feature fusion, the \( p' \) is repeated and upsampled to \( \mathbb{R}^{h \times w \times 3} \), then concatenated with the appropriately transformed tensor of the input image \( x \), resulting in a new tensor \( x' \in \mathbb{R}^{b \times h \times w \times (c+3)} \), which is passed through the Topology SegNet network for vessel segmentation. The segmentation result is denoted as \( y' \), which is compared with the ground truth \( y \). The model is trained using the combined loss function \( \alpha \cdot \mathcal{L}_{\text{Dice}} + (1 - \alpha) \cdot \mathcal{L}_{\text{CE}} \), where \( \mathcal{L}_{\text{Dice}} \) is the Dice loss and \( \mathcal{L}_{\text{CE}} \) is the cross-entropy loss. For simplicity, \( \alpha \) is set to 0.5. Meanwhile, the upsampled \( p' \) is also fed into the Topology SegNet, where subsequent feature fusion further enhances the segmentation performance and robustness of the model.

\noindent\textbf{PIs-Regressor.} As shown in Fig.~\ref{fig:fig1} (f), each pixel in the PI reflects the contribution of topological features to image. As mentioned in Preliminaries section, the ground-truth PI is obtained by applying Gaussian smoothing, which highlights global topological structures, such as vessel connectivity and circularity, not local details. This smoothing introduces inherent error tolerance, ensuring that even if there are local discrepancies in the predicted persistence image \( p' \), the overall topological features remain stable. For details, refer to Fig.~\ref{fig_2}. The encoder uses a pre-trained ResNet-50~\cite{he2016deep} to extract hierarchical features, and the decoder reconstructs the PI approximation using a DeConv Block, followed by a 1x1 convolution layer and Sigmoid activation to obtain \( p' \in \mathbb{R}^{b \times 1 \times h \times w} \). 

\noindent\textbf{Topology SegNet.} The fused image and topological features, denoted as \( x' \), are fed into our segmentation network, Topology SegNet, as illustrated in Fig.~\ref{fig_2}. Specifically, during each downsampling in Topology SegNet, the features produced by max-pooling are fused with the topological features, which are resized to match the dimensions of the downsampled feature map. The fused features are passed to the subsequent downsampling module for further processing. Further details are provided in Fig.~\ref{fig_2}. During upsampling, a similar operation is applied, except that the topological features are fused with the features obtained after deconvolution-based upsampling.

\section{Experiments and Results}
\label{sec:Experiments}

\noindent\textbf{Dataset.} Experiments are conducted on three curvilinear datasets: DRIVE \cite{staal2004ridge}, ER \cite{guo2021segmentation}, and a private ultra-widefield OPTOS retinal dataset from Xiamen Eye Center. Robustness is tested on perturbed DRIVE images (under/over-exposure, low contrast, blur).

\noindent\textbf{Evaluation Metrics.} 
Performance is measured by IoU, clDice~\cite{paetzold2019cldice}, and Dice coefficient and Betti Error~\cite{hu2019topology} using Betti numbers \( \beta_0, \beta_1 \) .

\noindent\textbf{Setting and Implementation.} Our network consists of two subnetworks with different learning rates and optimizers. 
PIs-Regressor adopts Adam (\(10^{-3}\)), while Topology SegNet uses SGD (\(10^{-2}\)) with a scheduler and 0.9 decay. 
Training runs for 500 epochs with a batch size of 4. 
Experiments are implemented in PyTorch, and Betti numbers are computed using GUDHI~\cite{maria2014gudhi}.

\begin{figure}[t]
    \centering
    \vspace{-6pt} 
    \includegraphics[width=\columnwidth]{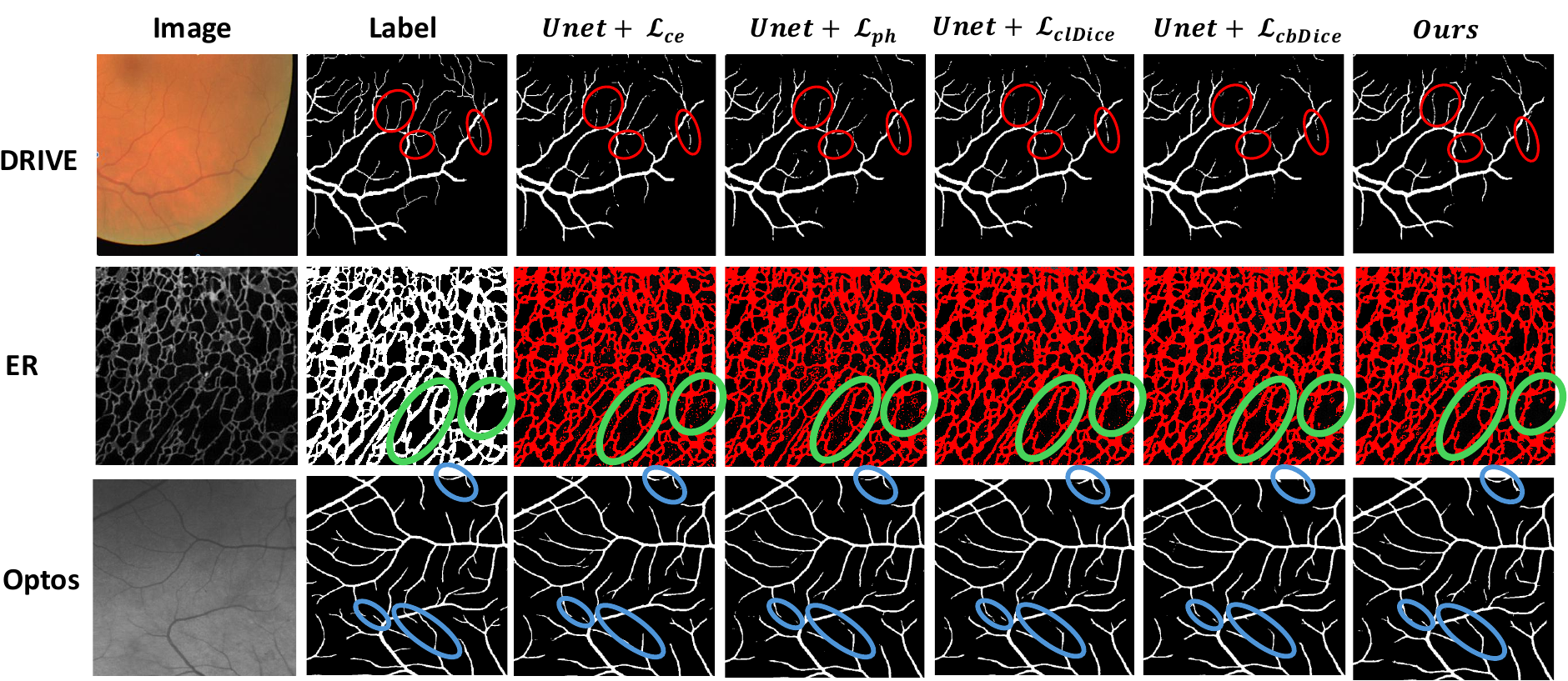}
    \caption{Visualization of segmentation results from different models on the DRIVE, ER, and Optos datasets.}
    \label{fig_3}
    \vspace{-6pt} 
\end{figure}

By integrating PI, our method significantly enhances both pixel accuracy and topological consistency in curvilinear segmentation (see Table 1). For instance, on the DRIVE dataset \cite{staal2004ridge}, our model with the traditional CE loss improves the Dice coefficient by 1.18 and reduces the topological error (measured by \( \beta_0 \)) by 90.4 compared to a standard U-Net with CE loss. These results indicate that the incorporation of topological features not only refines the pixel-level accuracy but also robustly preserves vessel connectivity, thereby mitigating issues such as vessel breaks. For the ER dataset \cite{guo2021segmentation}, the intricate morphology of the ER often leads segmentation models to lose structural continuity, resulting in isolated misclassified points and fragmented segments. The best-performing baseline is U-Net with clDice loss \cite{paetzold2019cldice}. In contrast, our model with clDice loss, which integrates PI, achieves a 2.88-point improvement in clDice along with significant reductions of 53.5 and 46.02 in \(\beta_0\) and \(\beta_1\), respectively. These improvements demonstrate that incorporating PI not only mitigates the occurrence of isolated missegmentations but also better preserves the continuous, cohesive structure of the ER. For more detailed visualizations, please refer to Fig.~\ref{fig_3}.

\begin{table}[t]
\caption{Evaluation on DRIVE, ER, and the Private dataset.}
\centering
\scriptsize 
\setlength{\tabcolsep}{2.5pt} 
\renewcommand{\arraystretch}{0.8} 
\resizebox{\columnwidth}{!}{ 
\begin{tabular}{|c|c|c|c|c|c|c|}
\hline
\multirow{2}{*}{\textbf{Dataset}} & \multirow{2}{*}{\textbf{Method}} & \multicolumn{5}{c|}{\textbf{Performance Metrics}} \\ \cline{3-7}
 &  & \textbf{Dice $\uparrow$} & \textbf{ClDice $\uparrow$} & \textbf{MIoU $\uparrow$} & \textbf{$\boldsymbol{\beta_0}\downarrow$} & \textbf{$\boldsymbol{\beta_1}\downarrow$} \\ \hline

\multirow{9}{*}{\textbf{DRIVE}} & CE Loss & 80.73 & 81.00 & 81.13 & 217.2 & 23.2 \\ 
\rowcolor[HTML]{EFEFEF} & Ours + CE Loss & 81.91 & 82.26 & 82.12 & 126.8 & 22.65 \\ 
& PH Loss~\cite{hu2019topology} & 81.62 & 81.83 & 81.89 & 148.05 & 21.7 \\ 
\rowcolor[HTML]{EFEFEF} & Ours + PH Loss~\cite{hu2019topology} & 81.83 & 82.06 & 82.10 & 123.6 & 25.1 \\ 
& clDice~\cite{paetzold2019cldice} & 81.76 & 82.17 & 81.97 & 115.05 & 26.2 \\ 
\rowcolor[HTML]{EFEFEF} & Ours + clDice~\cite{paetzold2019cldice} & 81.96 & 82.08 & 82.17 & 111.25 & 23.7 \\ 
& cbDice~\cite{shi2024centerline} & 81.76 & \textbf{82.57} & 81.95 & 108.55 & 23.3 \\ 
\rowcolor[HTML]{EFEFEF} & Ours + cbDice~\cite{shi2024centerline} & 82.00 & 81.82 & 82.26 & 116.45 & \textbf{20.95} \\ 
& \textbf{Ours+CE+Dice Loss} & \textbf{82.12} & 82.56 & \textbf{82.30} & \textbf{101.25} & 24.95 \\ \hline

\multirow{9}{*}{\textbf{ER}} & CE Loss & 82.45 & 86.65 & 77.49 & 414.08 & 32.8 \\ 
\rowcolor[HTML]{EFEFEF} & Ours + CE Loss & 83.76 & 91.16 & 79.34 & 83.58 & \textbf{29.13} \\ 
& PH Loss~\cite{hu2019topology} & 84.10 & 89.54 & 79.36 & 222.6 & 88.88 \\ 
\rowcolor[HTML]{EFEFEF} & Ours + PH Loss~\cite{hu2019topology} & 83.32 & 89.61 & 78.18 & 135.8 & 46.48 \\ 
& clDice~\cite{paetzold2019cldice} & 84.23 & 90.60 & 79.62 & 184.2 & 77.85 \\ 
\rowcolor[HTML]{EFEFEF} & Ours + clDice~\cite{paetzold2019cldice} & \textbf{84.52} & \textbf{93.48} & \textbf{79.80} & 30.70 & 31.83 \\ 
& cbDice~\cite{shi2024centerline} & 83.65 & 89.65 & 78.82 & 179.63 & 147.33 \\ 
\rowcolor[HTML]{EFEFEF} & Ours + cbDice~\cite{shi2024centerline} & 84.02 & 93.45 & 79.17 & \textbf{24.15} & 33.9 \\ 
& \textbf{Ours+CE+Dice Loss} & 82.73 & 91.44 & 76.80 & 35.68 & 54.88 \\ \hline

\multirow{9}{*}{\textbf{Private}} & CE Loss & 74.84 & 81.69 & 86.47 & 7.765 & 0.582 \\ 
\rowcolor[HTML]{EFEFEF} & Ours + CE Loss & 75.78 & \textbf{83.15} & 87.64 & 5.821 & 0.629 \\ 
& PH Loss~\cite{hu2019topology} & 74.58 & 81.36 & 86.74 & 7.083 & 0.549 \\ 
\rowcolor[HTML]{EFEFEF} & Ours + PH Loss~\cite{hu2019topology} & 75.54 & 82.65 & 87.52 & 6.869 & 0.534 \\ 
& clDice~\cite{paetzold2019cldice} & 74.89 & 81.85 & 86.97 & 7.906 & 0.587 \\ 
\rowcolor[HTML]{EFEFEF} & Ours + clDice~\cite{paetzold2019cldice} & 76.15 & 82.70 & 86.99 & 7.049 & 0.537 \\ 
& cbDice~\cite{shi2024centerline} & 75.15 & 81.08 & 86.80 & 8.503 & 6.010 \\ 
\rowcolor[HTML]{EFEFEF} & Ours + cbDice~\cite{shi2024centerline} & 76.43 & 81.67 & 87.14 & 4.410 & 7.289 \\ 
& \textbf{Ours+CE+Dice Loss} & \textbf{76.99} & 82.77 & \textbf{87.72} & \textbf{4.278} & \textbf{0.464} \\ \hline
\end{tabular}
}
\end{table}

 Among baseline methods, U-Net with cbDice loss~\cite{shi2024centerline} performed best on the private dataset. 
Using our model with CE and Dice loss improves Dice by 1.84, clDice by 1.69, and mIoU by 0.92, 
while reducing \( \beta_0 \) by 4.23 and \( \beta_1 \) by 5.55 compared to the best baseline. 
Integrating topological features without complex architectures or losses, our model enhances segmentation across domains 
and can be easily combined with other topology-based methods for further gains.

\noindent\textbf{Robustness Analysis.} The integration of PI reduces the impact of local image variations (e.g., blur, lighting, or contrast changes) on the segmentation network. Thus, integrating PI as a topological feature within the segmentation network can enhance robustness. A detailed analysis of the perturbed DRIVE \cite{staal2004ridge} dataset, as shown in Table 2, highlights that our model substantially improves robustness. Under four perturbations, traditional segmentation methods (UNet + $\mathcal{L}_{\text{ce}}$) show significant performance drops, particularly in blurred and overexposed conditions. Specifically, in the overexposed scenario, $\beta_0$ and $\beta_1$ increased significantly, by 1039.9 and 43.6, respectively, indicating more fragmentation and noise in the segmentation results. In contrast, our method saw a smaller increase in $\beta_0$ (76.9) and a decrease in $\beta_1$ (-0.8). Moreover, our model outperformed the best-performing comparison method, U-Net with cbDice, in Dice, clDice, and MIoU by 0.72, 0.06, and 0.68, respectively, demonstrating better connectivity and fewer fractures under overexposure.

\begin{table}[t]
\caption{Comparison of segmentation results on the DRIVE test set with four perturbations: Blur, Low Contrast, Underexposure, and Overexposure, using a model pretrained on the unperturbed DRIVE dataset.}
\centering
\scriptsize 
\setlength{\tabcolsep}{2.2pt} 
\renewcommand{\arraystretch}{0.85} 
\resizebox{\columnwidth}{!}{ 
\begin{tabular}{|c|c|c|c|c|c||c|c|c|c|c|}
\hline
\textbf{Method} & \multicolumn{5}{c||}{\textbf{Blur DRIVE}} & \multicolumn{5}{c|}{\textbf{Contrast DRIVE}} \\ \cline{2-11}
& Dice $\uparrow$ & ClDice $\uparrow$ & MIoU $\uparrow$ & $\boldsymbol{\beta_0}\downarrow$ & $\boldsymbol{\beta_1}\downarrow$
& Dice $\uparrow$ & ClDice $\uparrow$ & MIoU $\uparrow$ & $\boldsymbol{\beta_0}\downarrow$ & $\boldsymbol{\beta_1}\downarrow$ \\ \hline
CE Loss & 64.82 & 56.52 & 70.37 & 95.00 & 54.25 & 80.60 & 79.98 & 81.10 & 193.75 & 23.9 \\ 
PH Loss~\cite{hu2019topology} & 64.11 & 55.17 & 69.55 & 52.50 & 51.2 & 81.60 & 81.32 & 81.96 & 148.1 & 24.2 \\ 
clDice~\cite{paetzold2019cldice} & 64.90 & 56.70 & 69.89 & 64.25 & 50.5 & 81.93 & 81.82 & 82.18 & 122.6 & 25.8 \\ 
cbDice~\cite{shi2024centerline} & 62.85 & 52.93 & 68.44 & \textbf{32.2} & 49.9 & 81.89 & 81.95 & 82.14 & 111.1 & \textbf{24.0} \\ 
\rowcolor[HTML]{EFEFEF} \textbf{Ours+CE+Dice Loss} & \textbf{66.48} & \textbf{65.83} & \textbf{71.30} & 50.00 & \textbf{46.45} & \textbf{82.16} & \textbf{82.28} & \textbf{82.38} & \textbf{96.5} & 25.85 \\ \hline
\hline
\textbf{Method} & \multicolumn{5}{c||}{\textbf{Underexposed DRIVE}} & \multicolumn{5}{c|}{\textbf{Overexposed DRIVE}} \\ \cline{2-11}
& Dice $\uparrow$ & ClDice $\uparrow$ & MIoU $\uparrow$ & $\boldsymbol{\beta_0}\downarrow$ & $\boldsymbol{\beta_1}\downarrow$
& Dice $\uparrow$ & ClDice $\uparrow$ & MIoU $\uparrow$ & $\boldsymbol{\beta_0}\downarrow$ & $\boldsymbol{\beta_1}\downarrow$ \\ \hline
CE Loss & 78.86 & 77.87 & 79.80 & 150.80 & 29.55 & 69.39 & 64.20 & 70.83 & 1257.1 & 66.8 \\ 
PH Loss~\cite{hu2019topology} & 80.67 & 80.22 & 81.15 & 128.00 & \textbf{24.50} & 77.93 & 77.02 & 78.85 & 309.75 & 29.40 \\ 
clDice~\cite{paetzold2019cldice} & 80.88 & 80.26 & 81.32 & 118.55 & 27.55 & 74.72 & 72.86 & 75.79 & 324.40 & 65.05 \\ 
cbDice~\cite{shi2024centerline} & 80.86 & 80.46 & 81.30 & 101.5 & 24.85 & 77.93 & 77.90 & 78.65 & 207.05 & 34.30 \\ 
\rowcolor[HTML]{EFEFEF} \textbf{Ours+CE+Dice Loss} & \textbf{81.03} & \textbf{80.71} & \textbf{81.47} & \textbf{94.5} & 28.25 & \textbf{78.65} & \textbf{77.96} & \textbf{79.33} & \textbf{178.15} & \textbf{24.15} \\ \hline
\end{tabular}
}
\end{table}

\subsection{Conclusion}
In this study, we proposed an efficient approach for extracting topological features directly from the original images and integrating them into segmentation networks. The inclusion of topological features significantly enhances the model's performance, particularly in terms of robustness and vascular connectivity preservation, even under challenging conditions such as overexposure. Our work offers a novel perspective on designing a simple yet highly robust curvilinear segmentation model that shows promise in real-world medical scenarios.

\bibliographystyle{IEEEbib}
\bibliography{strings}

\end{document}